%% file: main_paper.tex
\documentclass[sigconf]{acmart}
\usepackage{url}
\usepackage{amsmath}
\usepackage{graphicx}
\usepackage[T1]{fontenc}

\setcopyright{none}

\newcommand{\secref}[1]{Section~\ref{sec:#1}}
\newcommand{\secstworef}[2]{Sections~\ref{sec:#1} and~\ref{sec:#2}}

\newcommand{\exref}[1]{Example~\ref{ex:#1}}

\newcommand{\appref}[1]{Appendix~\ref{app:#1}}

\newcommand{\figref}[1]{Figure~\ref{fig:#1}}
\newcommand{\figstworef}[2]{Figures~\ref{fig:#1} and~\ref{fig:#2}}

\newcommand{\tabref}[1]{Table~\ref{tab:#1}}

\newcommand{\clarozero}[0]{$0$} %
\newcommand{\claroone}[0]{$1$} %
\newcommand{\brevezero}[0]{$0$} %
\newcommand{\breveone}[0]{$1$} %

\renewcommand{\breve}[0]{BrevE}
\newcommand{\claro}[0]{CLaro}
\newcommand{\F}{F$_{1}$ }

\begin{document}
\title{A User-Centered Evaluation of Spanish Text Simplification}

\author{Adrian de Wynter}
\email{adewynter@microsoft.com}
\affiliation{%
    \institution{Microsoft Corporation}
    \streetaddress{1 Microsoft Way}
    \city{Redmond}
    \state{WA}
    \country{USA}
    \postcode{98052}
}
\author{Anthony Hevia} 
\email{anthonyhevia@microsoft.com}
\affiliation{%
    \institution{Microsoft Corporation}
    \streetaddress{1 Microsoft Way}
    \city{Redmond}
    \state{WA}
    \country{USA}
    \postcode{98052}
}
\author{Si-Qing Chen}
\email{sqchen@microsoft.com}
\affiliation{%
    \institution{Microsoft Corporation}
    \streetaddress{1 Microsoft Way}
    \city{Redmond}
    \state{WA}
    \country{USA}
    \postcode{98052}
}

\begin{abstract}
\input{abstract}
\end{abstract}

\maketitle

\input{main_content}

\bibliography{biblio}
\bibliographystyle{acl_natbib}

\end{document}

%% file: abstract.tex
We present an evaluation of text simplification (TS) in Spanish for a production system, by means of two corpora focused in both complex-sentence and complex-word identification. 
We compare the most prevalent Spanish-specific readability scores with neural networks, and show that the latter are consistently better at predicting user preferences regarding TS. 
As part of our analysis, we find that multilingual models underperform against equivalent Spanish-only models on the same task, yet all models focus too often on spurious statistical features, such as sentence length. 
We release the corpora in our evaluation to the broader community with the hopes of pushing forward the state-of-the-art in Spanish natural language processing. 

%% file: main_content.tex
\section{Introduction}\label{sec:intro}

Text simplification (TS) is a natural language processing task whose objective is to return a more readable version of the input text \cite{chandrasekar-etal-1996-motivations,Siddhartan,SaggionTS}. 
This may be achieved through simple, rule-based systems that (for example) replace words and idioms with shorter equivalents \cite{chandrasekar-etal-1996-motivations}. 
This simplification must preserve the semantic content of the original work. 
Improved readability via TS is often employed to make texts accessible in education \cite{Crossley2,Siddhartan} and for people with reading difficulties \cite{RelloDyslexiaFreq}. 

Contemporary TS emphasizes contextualization (that is, sentence-level simplification) \cite{gooding-kochmar-2019-complex,yimam-etal-2017-cwig3g2,AlvaManchego}, and models it as a machine translation task. 
In a production system such as Microsoft Word, it is desirable to focus on \emph{which} sentences could use simplification, as opposed to generating a potentially overwhelming number of rewrites. This approach (targeted simplification) drastically improves the quality of the rewrite \cite{paetzold-specia-2016-benchmarking}. 
It has the added benefit of reducing the number of calls to the model, but at the expense of making the TS system data-dependent. 
It makes sense to use a readability score to decide which sentences to send to the model, although automated measures tend to fail to fully capture user preference \cite{alva-manchego-etal-2021-un}.

In this paper we focus on Spanish TS. Spanish is spoken by over $471$M people in over $40$ countries, making it the second-most spoken first language in the world \cite{Ethnologue}. Unlike English, Spanish has a relatively lax syntax with freer word order thanks to inflection,\footnote{For example, Spanish has $48$ simple verb inflections, compared to English' 8 \cite{Whitley}.} and other grammatical properties that, from a user perspective, induces a nuanced variation in word choice and rewrite preferences \cite{Alarcos,Gutierrez} . %
We concentrate on two subtasks of Spanish TS: complex word identification (CWI) and complex sentence identification (CSI). 
We rephrase CWI as \emph{plain language identification}, or PLI, which emphasizes lexical simplification in-context, but preserves sentence structure. 
Analogously, CSI performs syntactic simplification at the sentence level. %
Note that these tasks are not complete overlaps of one another. Consider: 
\begin{align}
\text{Una enfermedad} \text{ originada por causas internas.}\nonumber\\
\text{\emph{(t. An illness originated by internal causes.)}}\label{ex:cwi}\\
\text{Una enfermedad}\text{ endógena.}\nonumber \\
\text{\emph{(t. An endogenous illness.)}}\label{ex:csi}
\end{align}

While \exref{csi} is arguably a syntactically simpler version of \exref{cwi}, the latter is lexically simpler. 
We discuss further examples of PLI and CSI in \secstworef{brevedata}{clarodata}.

\subsection{Contributions}
Our paper has three main contributions: 
\begin{enumerate}
\item We introduce and open-source\footnote{Available at \url{https://github.com/microsoft/BrevE-CLaro} along with its datasheet \cite{gebru2021datasheets}.} the corpora used for our analysis: \breve{} ("Corpus de la \underline{\textbf{Brev}}edad Linguística en \underline{\textbf{E}}spañol") for CSI, and \claro{} ("\underline{\textbf{C}}orpus del \underline{\textbf{L}}engu\underline{\textbf{a}}je Cla\underline{\textbf{r}}o en Españ\underline{\textbf{o}}l") for PLI.\footnote{Spanish for "The corpus of linguistic brevity in Spanish" and "The Spanish clear-language corpus". "Breve" and "claro" mean "concise" and "clear", respectively.} 

\item We find that common Spanish readability scores such as the Fern\'andez Huerta score \cite{FHScore}, Szigriszt \cite{Szigriszt}, and \textmu\ \cite{mumunoz,MnM} scores do not predict user preference well in CSI and PLI (\secref{experiments}) when compared to various Spanish and multilingual deep-learning models, and two large language models (LLM). 

\item We find that Spanish-only models outperform monolingual models in PLI, and match them in CSI. 
This is not in line with some findings \cite{yimam-etal-2017-multilingual,finnimore-etal-2019-strong}, although some evidence exists at the sentence-level scope \cite{vasquez-rodriguez-etal-2022-benchmark}. 

\end{enumerate}

\subsection{Related Work}\label{sec:relatedwork}

The most comprehensive work in TS is perhaps \citet{SaggionTS}, although the survey by \citet{AlvaManchego} is more up-to-date. We direct the interested reader to these works and the primers from \citet{Ludivine} and \citet{SiddharthanSurvey}. For readability scores, see \citet{Martinc}. See \citet{gooding-2022-ethical} for risks and harms of TS. 
Recent work has pivoted towards deep-learning based approaches, modeling it as a machine-translation (MT) task \cite{coster-kauchak-2011-simple,nisioi-etal-2017-exploring,surya-etal-2019-unsupervised,barzilay-elhadad-2003-sentence} with great success. 

\citet{stanjermeasures} found with statistical methods--not through user preference scores--that English readability scores are linearly correlated. 
\citet{azpiazu-pera-2019-multiattentive} found that neural networks outperformed them, but focused solely on words and not on other features. A similar conclusion was reached by \citet{Martinc}, who indicate that different models focus on different features. 
\citet{alva-manchego-etal-2021-un} showed that automated TS scores are inadequate at capturing user preference in English. 
Our results align with these findings: all of the models evaluated are better than the readability scores, none achieves above $85.5\%$ \F{} in PLI and $67.5\%$ \F{} in CSI, and we find them too focused on sentence length as opposed to sentence structure. 

\citet{vasquez-rodriguez-etal-2022-benchmark} introduced a neural benchmark for two-and-three-class text complexity. 
They showed that readability scores are unreliable over different text lengths, although the corpora they used for the evaluation was not human-annotated and mainly crawled from L2 learner resources. 
They also discuss Spanish readability scores, but do not evaluate newer scores such as the \textmu\, score. 
\citet{dePolini} and \citet{Crawoford} also present evaluation metrics, but are for elementary school texts and too narrow for our purposes. 
However, \citet{stajner-etal-2015-automatic} found that MT-based approaches outperformed rule-based approaches in Spanish based on user preference. 

TS corpora in Spanish are scarce. 
The best examples are both mined from news: the work by \citet{COLLADOS2013464}, a 3000-sentence parallel corpus, and Newsela \cite{xu-etal-2015-problems} (250 articles with varying simplification levels). 
CWI is considered a special type of TS, with datasets and solutions designed for it \cite{yimam-etal-2017-cwig3g2}. 
For Spanish CWI, \citet{bott-etal-2012-spanish} built a system for simplification that computes simplicity based on both the word length and frequency in a corpus. 
\citet{SAGGION2016200} built a similar system based on lexical simplification with contextualization. 
Our aim is not to build a parallel corpus, but we leverage some of the rules found by \citet{bott-etal-2012-spanish}. %

Recently \citet{yimam-etal-2017-multilingual} released a sentence-based corpus for CWI. 
We consider our work to be complementary to theirs, and include the learnings from \citet{yimam-etal-2017-cwig3g2,finnimore-etal-2019-strong,yuan2021synthbio}: namely how to ensure quality, high-volume annotations. %

\section{Datasets}\label{sec:datasets}

Our data creation process is similar to the one by \citet{yuan2021synthbio}. 
It involved gathering sentences, simplifying them with LLMs, and then requesting human annotators to provide a better rewrite. 

We used two source datasets: 
CWI 2018 \cite{yimam-etal-2017-multilingual,yimam-etal-2017-cwig3g2}, and OSCAR's February 2021 version \cite{oscar3,oscar4}. 
Although CWI 2018 is pre-annotated, it is small and sourced from Wikipedia, making it too skewed from a generalizability standpoint. 
It has also been pointed out that Wikipedia too skewed and not fully effective for TS \cite{xu-etal-2015-problems}. 
OSCAR contains informal and conversational sentences, but requires manual intervention to ensure responsible AI (RAI) practices.

\subsection{CWI 2018 Corpus}

CWI 2018 is sampled from Wikipedia and is meant for sentence-based lexical simplification. 
We focus on the Spanish split, comprised of $1,387$ sentences annotated by crowdsourced native speakers. 
The annotations are one-to-many, as the annotators were tasked to highlight and rate words and phrases from a given sentence. \cite{yimam-etal-2017-multilingual}. 
The authors attribute the low inter-annotator agreement of this dataset to multiple phrases being selected. 
This learning is our motivation for rephrasing CWI as PLI. 

We selected from CWI 2018 all unique source sentences, and filtered out these that were too short, lacking enough context, or too hard to correct grammatically. 
Likewise, we removed repeated sentences with common structures such as "$\langle$Person$\rangle$  ($\langle$date$\rangle$) was a $\langle$occupation$\rangle$" or "$\langle$Toponym$\rangle$ is part of $\langle$toponym$\rangle$". 
The final number of selected sentences was $736$. 
To maintain the integrity of the original task, we did not use the test set for CWI 2018. %

\subsection{OSCAR}

OSCAR is a very large corpus sampled from the Common Crawl,\footnote{\url{https://commoncrawl.org/}} stripped from web content and split into languages. 
The raw corpus contains a significant proportion of sentence fragments, paragraphs, and offensive content. 
Prior to sampling from the Spanish split from OSCAR, we deduplicated and filtered the data based on a list of words and sentences that could cause harm. 
We manually cleaned the sentences from symbols and tags not related to the discourse. %

We sampled sentences from OSCAR based on a list of uncommon and difficult words as per considered by native speakers. 
This list was harvested based on multiple lists of complex words crawled through the internet. It totaled about $800$ words ($1,200$ after expanding based on morphological properties such as gender). 
The list was further cleaned by removing non-Spanish sentences and technical terms. 
The only exception to this last rule was when these words had a plain-language equivalent ("hipercolesterolemia" = "(con) alto colesterol"). 
We kept multi-syllable examples that weren't complex words (e.g., "francachela") as adversarial examples, but verified their use with the RAE dictionary.\footnote{An authority on the Spanish language: \url{https://dle.rae.es/}} 

We selected sentences that emphasized localisms: for example, sentences containing voseo and associated conjugations; or sentences containing frequent Nahuatl terms (e.g., "tianguis"). 
OSCAR contributed $4,000$ sentences to our corpus.

\subsection{Dataset Construction}

We generated "pre-simplified" versions of the sentences with two models: GPT-3 \cite{GPT3}, prompted to simplify the input sentence; and then with a proprietary TS seq2seq transformer model based on Turing.\footnote{\url{https://turing.microsoft.com/}} %
We kept the sentences as an aligned corpus, even in the case where no simplification could be found by the models. 

\subsection{Annotation}\label{sec:annotation}
The aligned corpus was annotated by five professional annotators, all native Spanish speakers: two identifying as male and three as female. 
The annotators have training in linguistics and were chosen based on their original dialect, as well as their familiarity with other variants. 
They are native in Castilian (as spoken in Central-Northern Spain), Venezuelan, Argentine, and Mexican Spanish; and familiar with Colombian, Chilean, Peruvian, Dominican, Puerto Rican, Honduran, and Cuban Spanish. 
Note that native competence influences the ability of the annotators to perform CWI \cite{paetzold-specia-2016-semeval}. 
They were contracted through an annotation services company and remunerated for their work at a rate starting at $\$20$ USD/hr.

The annotators were given pairs of the form $(\text{source}, \text{target})$, and asked the following:\footnote{The full rubric also evaluated semantic loss through simplification \cite{devaraj-etal-2022-evaluating}, requesting a rewrite in this case. It is in \url{https://github.com/microsoft/BrevE-CLaro/rubric.pdf}.} 
\begin{enumerate}
    \item Does the source need simplification?
    \item Do you prefer the source, or the target?
    \item Are there any grammatical errors in the target?
    \item Is there a simpler way to write this sentence without splitting it into its constituent parts?
    \item Is there any content in this pair that may cause harm or exclusion to someone; or be considered offensive?
\end{enumerate}
For our paper we measured the inter-annotator agreement for the second question with Fleiss' kappa: $\kappa = 87\%$. 
We used the responses to questions 3, 4 and 5 to update (or drop, in the case of question 5) sentences in our aligned corpus, with at least a $60\%$ agreement on the suggested rewrites required for an update. 

\subsection{Responsible AI}\label{sec:rai}
To ensure that the datasets would be used responsibly after their release (and in addition to the filter mentioned) the annotators were requested to flag problematic content (question 5 in \secref{annotation}). 
In the entire corpus, $14$ sentences were flagged by at least one annotator. We removed these sentences from the dataset. 
The entire corpus was scanned with a named entity recognizer, and all names of existing people were replaced by randomly assigned names.\footnote{The random assignment is sentence-level: names appearing more than once in a sentence pair are replaced by the same string.} 

\subsection{\breve}\label{sec:brevedata}

\breve{} was built by randomly selecting either the source or the target in the aligned, updated corpus. Sources are considered needing simplification by the annotators (label: $1$), based on a minimum agreement of $80\%$. Otherwise, it would be marked as $0$. %

We manually selected counterexamples based on length (e.g., short sentences labeled as \breveone) and added them to the test set. 
We also removed sentences longer than $150$ words. 
The final size of \breve{} is $3,729$ sentences, separated in (train/test) $2,929/800$. 
Samples of this corpus are in \tabref{brevesamples}. 
Plots depicting the length and lemma distributions are shown in \secref{experiments}. 

\begin{center}
\begin{table}[h]
\begin{tabular}{| p{0.75\columnwidth} | c |} \hline
Source & Label  \\ \hline
Son los adecuados, de forma general, para situaciones en la que existe muy poca luz ambiente. & 1 \\ \hline
Al final del mismo resultó ganadora la pareja formada por Scott y Karim. & 1 \\ \hline
Sin pensarlo pidió el temulento una botella de licor que nunca se acabase. & 0 \\ \hline
Hacerse el lipendi es a veces necesario, pero Don Sandro se pasó tres pueblos. & 0 \\ \hline
\end{tabular}
\caption{Sample entries for \breve{}. The first two sentences have syntactical simplifications, but the last two do not. 
The third line does have a lexical simplification ("temulento", an uncommon word) which alters its label in \claro{}. 
The second entry is more formal than the fourth, which uses colloquialisms and metaphors (t. "\emph{pretending to be a nit is sometimes necessary, but Don Sandro went three towns over the line.}"). 
"Tres pueblos" ("\emph{three towns}") could be omitted, but it is unclear if it is literal or figurative, and hence it is left in to avoid loss of meaning.}\label{tab:brevesamples}
\end{table}
\end{center}

\subsection{\claro}\label{sec:clarodata}

To build \claro{} we expanded the list from our aligned dataset with the one by \citet{bott-etal-2012-spanish}, plus our own list of phrasal mappings to their simpler language equivalents (e.g., "aquel" to "ese", "nos dirigimos" to "vamos"). 
We substituted them on the target sentences when the simplification in \breve{} had matches. 
We re-ran annotation to verify labelling, and removed all sentences where agreement was below $60\%$. 
Grammaticality was verified by a single annotator. 

In \claro{}, sentences with label=$1$ are those that can be rewritten in a simpler way; whose meaning cannot be inferred from the context; 
or that can be rewritten with a reduced lexicon. To account for the wide range and variation of Spanish dialects, colloquialisms aren't considered complex language. %

We created the splits following \secref{brevedata}, but we did not perform any manual selection of sentence pairs. 
Instead, we kept the same test set as \breve{}, but with probability $1/2$ we flipped the label and replaced it with the corresponding source (r. target) from the back-substituted corpus. 
Note that these are \emph{not} noisy labels, but solely ensuring disjoint source and targets from \breve. 
For comparison purposes, we retain the same label ratio, with a slight (0.8\%) discrepancy in the proportion of label: $0$. 

The final size of \claro{} is $3626$ sentences, split in (train/test) $2826/800$. The final overlap with \breve{} is $13.5\%$, or $108$ sentences, in the test set. 
Samples of this corpus are in \tabref{clarosamples}. 
See \secref{frequency} for a lexical distribution of the labels. 

\begin{center}
\begin{table}[h]
\begin{tabular}{| p{0.75\columnwidth} | c |} \hline
Source & Label  \\ \hline
La fauna autóctona la forman guazunchos, iguanas, y zorros. & 1 \\ \hline
El atolón se localiza unos 800 kilómetros al sur del Ecuador. & 1 \\ \hline
De vuelta al océano, por un descuido, la máscara de buceo se le cayó en un hoyo. & 0 \\ \hline
En la cocina mexicana de hoy, el mole suele acompañar carnes cocidas. & 0 \\ \hline
\end{tabular}
\caption{Sample \claro{} entries. The first two have phrasal syntactical simplifications that fall into the PLI paradigm. 
In the fourth line, "suele acompañar" (\emph{"it often accompanies"}) has a syntactical simplification that makes the sentence more concise ("acompaña"; t. \emph{"accompanies"}) but not a meaning-preserving lexical simplification to make the language clearer. 
The first line is syntactically simple, but has lexical simplifications ("autóctona"; t. \emph{"autochthonous"}).}\label{tab:clarosamples}
\end{table}
\end{center}

\section{Experiments}\label{sec:experiments}

In this section we evaluate and compare \breve{} and \claro{} as they relate to user preference and readability scores. 
We describe the readability scores (\secref{measures}) and then show our results when compared with variations of the BERT \cite{BERT} and DistilBERT \cite{DistilBERT} architectures (\secref{measurecomparison}). 
We also report our results on generalizability between PLI and CSI (\secref{crossstudy}), and the impact of word length, frequency, and morphology (\secstworef{erroranalysis}{frequency}). 

\subsection{Readability Scores}\label{sec:measures}

The Fernández Huerta score is a Spanish-tuned version of the Flesch-Kincaid score \cite{FKScore}. 
In our work we evaluated the corrected version, as per \citet{Gwillim}, that scales well to longer contexts. 
We denote it as $FH$. 
The Szigriszt score is an adapted version of the Fernández Huerta score with a new scale for interpretability. 
We denote it as $SP$. 
The \textmu\ score accounts for statistical measures of the text. 

All scores are lexical and rely on counting syllables and word frequencies. 
Let $s_w, w_\sigma$ be the mean number of syllables per word, and the mean number of words per sentence, respectively. Then if $mean(\hat{w})$ and $var(\hat{w})$ are the mean and variance of the word lengths $\hat{w}$ of a text with a total of $n$ words, the measures are: 

\begin{align}
    FH &= 206.84 - 60\cdot s_w - 102\cdot\frac{1}{w_\sigma} \\
    SP &= 207 - 62.3\cdot s_w - w_\sigma \\
    \mu&= \left(\frac{n}{n - 1}\right)\cdot\left(\frac{mean(\hat{w})}{var(\hat{w})}\right)\cdot100.
\end{align}

All these scales rate the reading ease between 0 (hardest) and 100 (easiest), although the threshold to judge the difference between, say, hard and average varies across them \cite{Barrio,mumunoz}. 

\subsection{Readability Comparison}\label{sec:measurecomparison}

We compared readability scores and neural networks when predicting user preference in PLI and CSI, as described by \breve{} and \claro{}. 
We evaluated two types of neural networks: 
\begin{enumerate}
    \item \emph{Monolingual}: BETO \cite{CaneteCFP2020} and DistilBETO \cite{canete-etal-2022-albeto}. They are the Spanish-only versions of BERT and DistilBERT. 
    \item \emph{Multilingual}: mBERT \cite{BERT} (multilingual BERT), a multilingual version of DistilBERT, and XLM-R \cite{XLMRoBERTa}. This last model was the highest-performing multilingual model publicly available for Spanish.\footnote{For sentence classification in XTREME \cite{XTREME} at the time of writing this.} 
\end{enumerate}

\begin{figure}[h]
\centering
\includegraphics[width=\columnwidth]{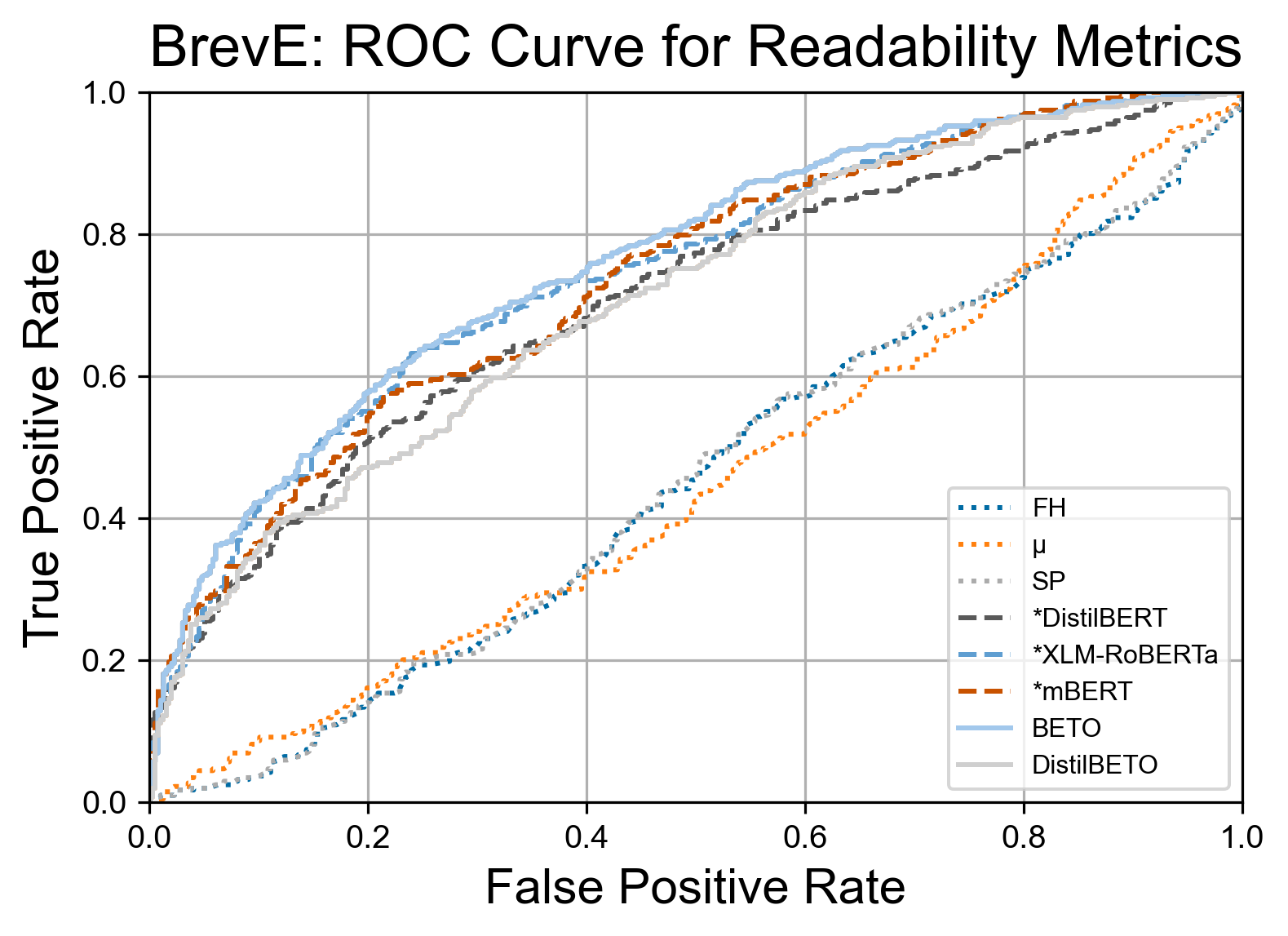}
\caption{Comparison of readability scores (dotted lines) and neural networks for \breve{}. 
There is no noticeable gap in performance between monolingual (solid lines) and multilingual (dashed lines, with an asterisk) models. All neural networks outperformed the readability scores.}\label{fig:brevefig}
\end{figure}

The ROC curves for our analysis are in \figstworef{brevefig}{clarofig}. 
Full results with scores are in \appref{evalbreakdown}, including an LLM evaluation. 
Overall we observed a gap between readability scores and neural networks. 
We also observed a gap between monolingual and multilingual models, particularly in \claro{}. 

\begin{figure}[h]
\centering
\includegraphics[width=\columnwidth]{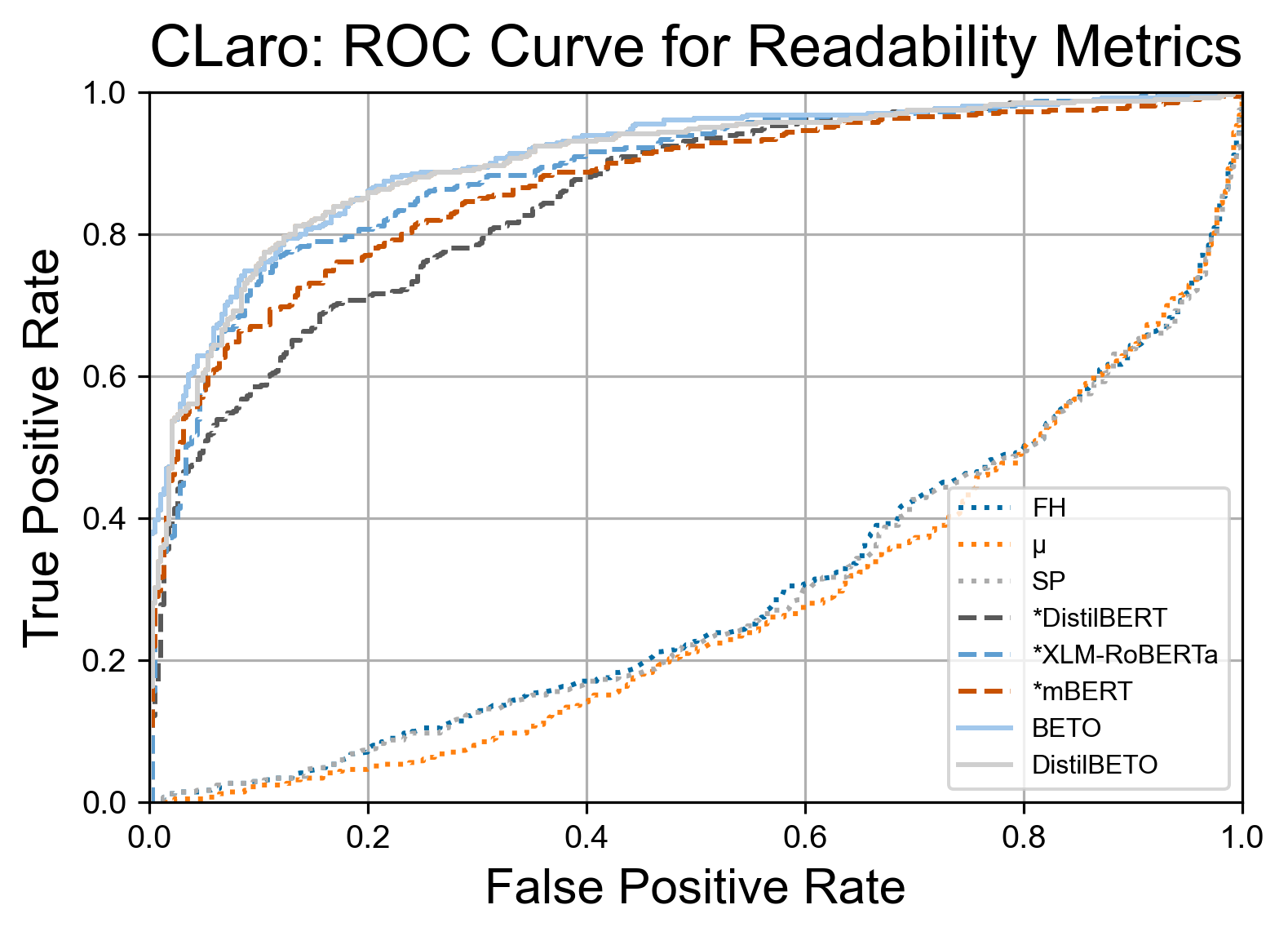}
\caption{Comparison of readability scores and neural networks in \claro{}. 
There is a gap between monolingual (solid lines) and multilingual (dashed, with an asterisk) models, with the former generally performing better. 
The largest multilingual model tested, XLM-R, had similar performance as the smaller monolingual models.}\label{fig:clarofig}
\end{figure}

\subsection{Cross-Corpora Studies}\label{sec:crossstudy}

We performed a cross-study where the models are trained on one task (say, \breve{}) and evaluated on the other (r. \claro{}). 
The performance of the models in this scenario drops noticeably: on average, of 14.1\% \F{} for training in \breve{} and testing in \claro{}; and 17.3\% \F{} for training in \claro{} and testing in \breve{}. 
In the first scenario, XLM-R had the largest drop (19.0\% \F{}) and DistilBETO the lowest (8.9\%). 
In the second, DistilBETO had the largest (19.8\% \F{}), and DistilBERT the lowest (14.9\%). 

\subsection{Error Analysis}\label{sec:erroranalysis}

\begin{figure}[h]
\centering
\includegraphics[width=\columnwidth]{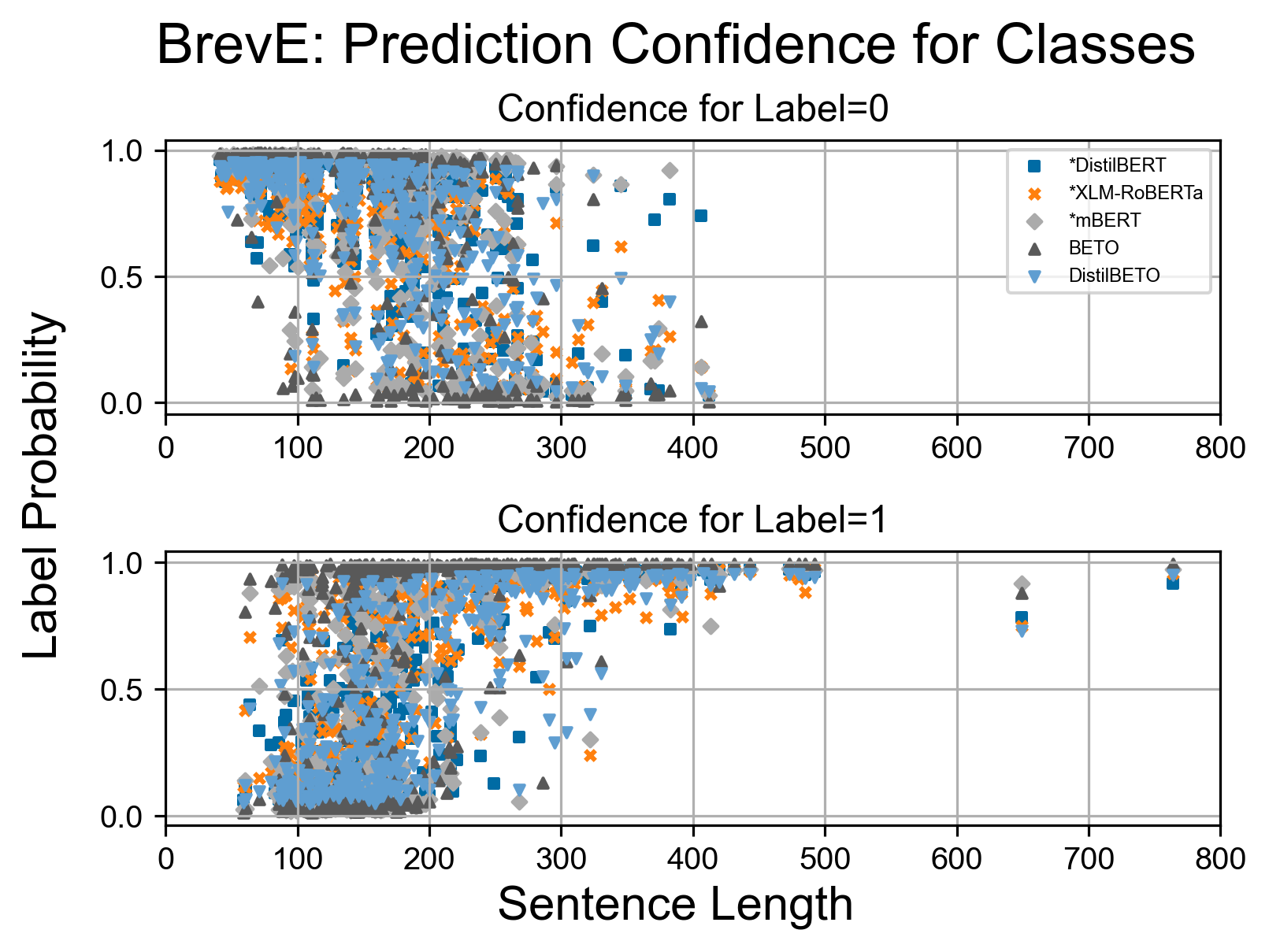}
\caption{Label confidence for models in \breve{}. Multilingual models are marked with (*). 
Models were more confident when the sentence length was $>200$, and there was little agreement on lower-length sentences. 
This suggests that they were too focused on sentence length as a feature, instead of the correct grammatical features.
}\label{fig:brevelen}
\end{figure}

\begin{figure}[h]
\centering
\includegraphics[width=\columnwidth]{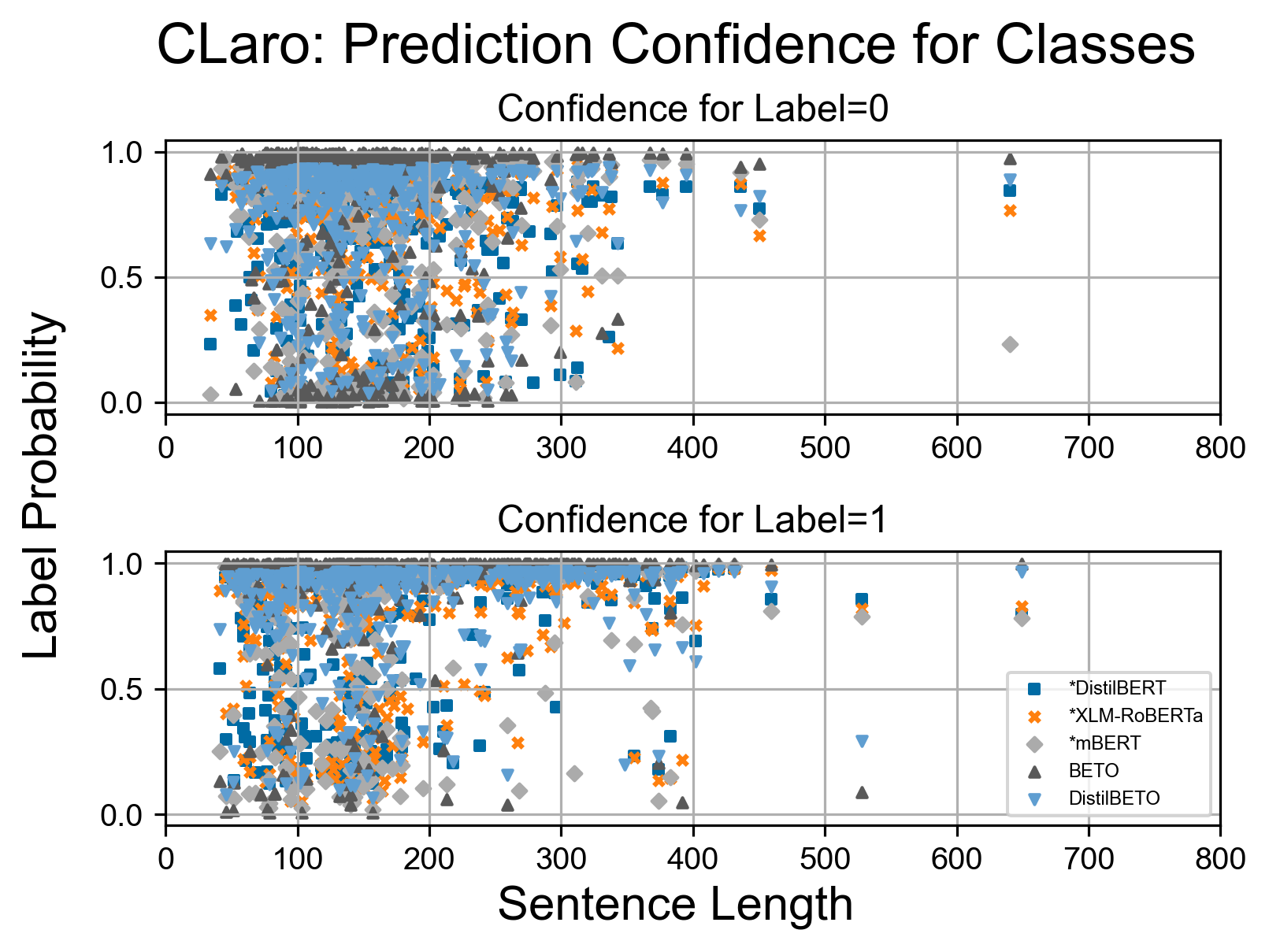}
\caption{Label confidence for models in \claro{}. Multilingual models are marked with (*). 
The best models, like DistilBETO (cyan $\blacktriangledown$), have high confidence for each label independent of sentence length. This suggests that it focused less on this feature, and learned the right grammatical structures.}\label{fig:clarolen}
\end{figure}

We found that most models learned a cutoff for the labels based on length, of around $220$ characters in \breve{} (\figref{brevelen}) and $180$ in \claro{} (\figref{clarolen}). 
The models that were the most successful (e.g., DistilBETO in \claro{}) showed relatively fewer instances of low-confidence samples at short sentence lengths. 

\subsection{Word Frequency and Representativeness}\label{sec:frequency}

In this analysis we evaluated the difference of PLI and CSI as it concerns to the lexicon. 
This was motivated by our observation that the annotators marked complex words based on the stem. 
For example, the Spanish word for journalism, "periodismo", would have long-syllable inflections such as "periodísticamente" or "periodístico", and neither were marked as complex. 
For this we lemmatized \breve{} and \claro{} using Stanza \cite{qi2020stanza}. 
We show in \figstworef{brevedistplots}{clarodistplots} the word distribution relative to the positive and negative labels. 

We found that \breve{} as a syntactic simplification task draws the lemma frequency curve down, and the tail up, thus lowering the overall statistical complexity of the corpus. 
This is supported by the fact that label=\brevezero{} contains $9,535$ lemmas, versus label=\breveone{}'s $9,101$. 
\claro{} props high-frequency lemmas up and clips the tail, in line with what would be expected from lexical simplification. 
Although label=\claroone{} for both datasets has the same lemmas, label=\clarozero{} has $6,906$ lemmas, making it a statistically easier task.

\begin{figure}[h]
\centering
\includegraphics[width=\columnwidth]{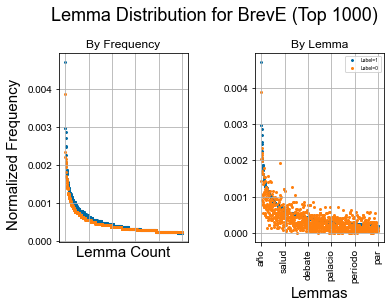}
\caption{Lemma distributions for labels in \breve{}. 
\emph{Left}: Lemma frequency comparison of both labels based on sorted count. 
\emph{Right}: Lemma-by-lemma comparison between sorted (label=\breveone{}) and their frequency in label=\brevezero{}. 
In \breve{} the simplification draws the frequency curve down, and brings the tail up, squishing the distribution and lowering the statistical complexity: see the rightmost plot, where several high-frequency lemmas have been zeroed. 
}
\end{figure}\label{fig:brevedistplots}

\begin{figure}[h]
\centering
\includegraphics[width=\columnwidth]{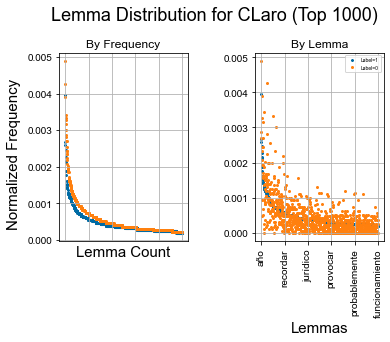}
\caption{Lemma distributions for labels in \claro{}. 
\emph{Left}: Lemma frequency comparison of both labels based on sorted count. 
\emph{Right}: Lemma-by-lemma comparison between sorted (label=\claroone{}) and their frequency in label=\clarozero{}. 
In \claro{} the simplification draws the frequency curve up, making high-frequency tokens more frequent and clipping the tail. 
In the rightmost plot observe that although some lemmas are zeroed, several high-frequency lemmas are now more common.
}
\end{figure}\label{fig:clarodistplots}

\subsection{Discussion}\label{sec:discussion}

Our error analysis suggests that most models tend to focus on spurious, lexical features. 
Regardless, they were able to outperform readability scores with ease. It is possible that our models have been biased based on the training distribution, as evidenced by our cross-study. 
Alternatively, these results show that the models are learning correctly the distinction between PLI and CSI, but the tasks are disjoint enough for the models to not be able to generalize to one another by only seeing one distribution--and hence are separate tasks. 
This is supported by our frequency analysis, where we observed that frequency-based replacement is a reasonable approach for PLI, but not CSI. 
Successful approaches to \breve{} could require the ability to capture linguistic relations beyond lemma frequency. 

\section{Limitations}\label{sec:limitations}

The first limitation of our work is technical. 
English-based TS measures have moved away from rule-based models and into neural approaches \cite{AlvaManchego}, such as SARI \cite{SARI} and SAMSA \cite{SAMSA}. 
They rely on a pretrained model (e.g. BERT) for parsing the model and correlate well with user preferences. 
SARI requires multiple references, and we only consider one. 
This is by design, however: our datasets involve identification, not simplification, of the source text. 
We did not evaluate SAMSA: it requires a word aligner, and we were unable to find quality aligners for Spanish. 

The second limitation is related to our user base. 
TS is a task that strongly depends on user preference. 
As indicated by the annotators, this is more evident in Spanish, due to its morphology and dialect diversity. 
While this would make simplification, not identification (the focus of this paper), more difficult, it is still worth remembering when deciding whether something needs simplification or not. 

We made an attempt to maintain representation and keep dialectal differences intact, but we were unable to involve representatives of all dialects. 
This means that our work is not properly localized. 
Without it, measures of PLI and CSI can only be an approximation. 
For example, the word "faramallero" is common in the Southern Cone, but not in Mexico, where it would be considered obscure.

\section{Conclusion and further work}\label{sec:conclusion}

In this paper we introduced two corpora and compared them to readability scores for TS in Spanish, for PLI (formerly CWI) and CSI. 
We found that the latter are not as effective as neural networks at capturing user preferences in either task. 
Our comparison of monolingual and multilingual models showed the latter to underperform Spanish-only models in PLI and CSI; and our error analysis showed that most models focused too much on spurious features, such as sentence length, and not enough on linguistic features. 
The results from our cross-study and our lemma analysis justifies distinguishing PLI from CSI, and makes them separate tasks with their own solution strategies. 

Our work can be expanded in a number of ways based on that: the models we evaluated can learn grammatical features with ease \cite{bertology}, so it could be a training issue. 
Most LLMs were unable to outperform models as small as DistilBETO, which suggests their linguistic capabilities are heavily focused in English. 

We did not evaluate automated benchmarks, such as SAMSA and SARI; or extended metrics such as NANO \cite{NANO} to Spanish. 
The latter could be a great addition to Spanish NLP: our work emphasized Spanish diversity, but is not fully dialect-centered. Further work could be done in creating a \emph{localized} TS dataset. %

\breve{} and \claro{} are not MT datasets, but classification datasets: we measure \emph{identification}, rather than \emph{simplification}. 
This makes our work only applicable to parts of TS, but further work could reconstruct the source or targets from our corpora to generate aligned corpora. 

\appendix

\section{Evaluation Breakdown Results}\label{app:evalbreakdown}

In \tabref{allresults} we display the performance of the models in our paper, plus 
two LLMs: GPT-3 and GPT-4 \cite{GPT4}. %
Both evaluations followed the author's guidelines for GPT-3, with a $50$-shot example and prompting the model by asking it to answer only $0$ or $1$. 
We ran this experiment five times, and report the highest-performing result. 
There is no indication that GPT-3 was intended to be used in Spanish, which explains the scores. 
On the other hand, GPT-4--an explicitly multilingual LLM--fared better, but did not outperform the other, more specialized models.   

\begin{center}
\begin{table}[h]
\begin{tabular}{| c || c | c || c | c |} 
 \hline
 Model &  \breve{} Ac. & \breve{} \F{} & \claro{} Ac. & \claro{} \F{} \\ \hline\hline
BETO                & $\mathbf{69.12}\%$ & $67.20\%$ & $83.40\%$ & $84.87\%$ \\ \hline
DistilBETO          & $63.38\%$ & $56.59\%$ & $\mathbf{84.52\%}$ & $\mathbf{85.48\%}$ \\ \hline\hline
*mBERT & $67.62\%$ & $63.88\%$ & $80.92\%$ & $80.91\%$ \\ \hline
*DistilBERT & $64.12\%$ & $56.45\%$ & $77.15\%$ & $77.76\%$ \\ \hline
*XLM-R & $\mathbf{69.12}\%$ & $\mathbf{67.46}\%$ & $82.02\%$ & $82.44\%$ \\ \hline\hline
**GPT-3      & $58.00\%$ & $39.78\%$  & $43.95\%$ & $56.05\%$ \\ \hline
*GPT-4      & $59.38\%$ & $56.49\%$  & $67.42\%$ & $67.66\%$ \\ \hline
\end{tabular}
\caption{Breakdown for our results on \breve{} and \claro{}. 
Models with an asterisk (*) are multilingual, and with two asterisks (**) English-only models. 
In \claro{} we saw a large gap between the Spanish-only and the multilingual models. %
In \breve{}, XLM-R performed about the same as BETO, and underperformed BETO and DistilBETO in \claro{}.}\label{tab:allresults}
\end{table}
\end{center}